\documentclass[letterpaper, 10 pt, conference]{ieeeconf}  

\IEEEoverridecommandlockouts                              

\usepackage{amsmath} 
\usepackage{amssymb}  
\usepackage{lipsum}
\usepackage{booktabs}
\usepackage{algorithm}
\usepackage{algpseudocode}
\usepackage{setspace}
\usepackage{dblfloatfix}
\usepackage{graphicx}
\usepackage{multirow}
\usepackage{lipsum}
\usepackage{tabularx}
\newcolumntype{P}[1]{>{\centering\arraybackslash}p{#1}}
\newcolumntype{M}[1]{>{\centering\arraybackslash}m{#1}}
\newcolumntype{C}[1]{>{\centering\arraybackslash}m{#1}}

\algrenewcommand\algorithmicindent{0.6em}

\algnewcommand{\CustomComment}[1]{ \hspace*{37.5mm}{#1}}
\usepackage{subcaption}
\usepackage{hyperref}
\usepackage{overpic}
\usepackage{xcolor}
\usepackage{bm}
  \newcommand{\panellabel}[1]{%
    {\setlength{\fboxsep}{1.2pt}\colorbox{white}{\color{black}\scriptsize #1}}}

\newcommand{\timelabel}[1]{
\put(1.5,65.17){\makebox[0pt][l]{\raisebox{-\height}{\panellabel{#1}}}}}
\newcommand{\kstlabel}[1]{
\put(98.5,1.5){\makebox[0pt][r]{\raisebox{\depth}{\panellabel{#1}}}}}

\graphicspath{{images/}}
\newlength{\panelheight}
\title{{\fontsize{15.9}{18}\selectfont\bfseries Imp-ACT: Adaptive Impedance Control and Action Chunking with Transformers to Learn Contact-Rich Manipulation from Demonstrations}}
\author{%
Luca Zanetti$^{1}$, Doganay Sirintuna$^{1}$, Idil Ozdamar$^{1}$, Pietro Balatti$^{1}$, Heng Zhang$^{2}$, Arash Ajoudani$^{1}$%
\thanks{$^{1}$\,\emph{Human-Robot Interfaces and Interaction Laboratory, Istituto Italiano di
Tecnologia, Genoa, Italy} Email: \texttt{first.last@iit.it}}%
\thanks{$^{2}$\,\emph{Safe AI Lab, Carnegie Mellon University} Email: \texttt{hengzhang01@cmu.edu}}%
}

\begin{document}

\maketitle
\thispagestyle{empty}
\pagestyle{empty}

\begin{abstract}

Contact-rich manipulation requires robots to balance accurate motion tracking with compliant interaction, yet most visual-action policies leave compliance fixed at the controller level. We present Imp-ACT, a methodologically grounded and practical approach to incorporating direction-dependent Cartesian stiffness modulation directly into demonstration collection, without manual stiffness selection or offline target reconstruction. During teleoperation, a self-tuning impedance controller adapts stiffness along the instantaneous direction of motion while maintaining compliance in orthogonal directions. The adapted stiffness is applied and recorded alongside visual observations and motion commands, capturing motion and compliance under the same dynamics. We implement this pipeline using Action Chunking with Transformer (ACT) to predict end-effector pose, gripper action, and motion-direction stiffness from visual, proprioceptive, and wrench observations. The performance of Imp-ACT is evaluated on wiping and plug insertion using both success rate and quantitative measures of contact behavior. Compared with fixed low- and high-stiffness baselines, Imp-ACT achieves comparable or higher success while maintaining low interaction forces. In wiping, it reduces contact-force vibration by approximately $29\times$ relative to the compliant baseline and $180\times$ relative to the stiff baseline. In plug insertion, it reduces forces orthogonal to the insertion direction by $43\%$ relative to the better fixed-stiffness baseline. These results highlight the benefit of maintaining sufficient stiffness along the direction needed for task execution while preserving compliance in other directions to limit contact forces and accommodate environmental constraints.

\end{abstract}


\section{Introduction}

Traditionally, robots have been deployed in structured environments where tasks and workspace conditions are known in advance. These controlled settings allow robot behaviors to be explicitly programmed, with limited need for adaptation because operating conditions remain largely unchanged \cite{VILLANI2018248}. As robotic applications extend beyond such predictable environments, manually programming behaviors for different situations becomes increasingly challenging. Imitation learning offers an alternative by enabling robots to acquire manipulation skills from expert demonstrations rather than explicitly specifying the behavior for every possible task condition \cite{lfd_review}.
Recent advances in deep learning have considerably improved the ability of imitation learning methods, such as ACT \cite{ACT} and Diffusion Policy \cite{chi2024diffusionpolicy}, to learn visuomotor behaviors from relatively small sets of demonstrations. However, most existing policies formulate the robot action primarily as a motion command, such as joint positions or end-effector poses. For contact-rich manipulation, motion alone does not fully determine the resulting behavior: task execution also depends on how the robot responds to interaction forces.

\begin{figure}[t]
  \centering
\includegraphics[width=0.85\linewidth]{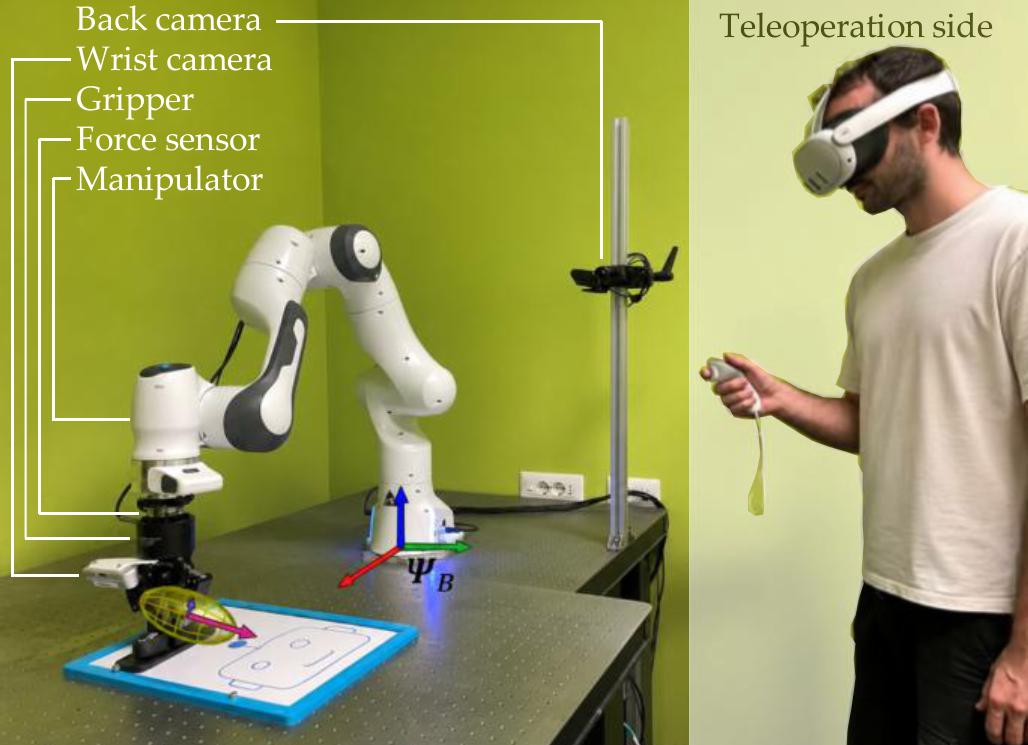}
  \caption{Imp-ACT turns intuitive human demonstrations into compliant, adaptive robot skills. By learning from human motion while an online Cartesian impedance controller generates direction-dependent stiffness, it unites natural kinematic teaching with interaction-aware control.
}
\label{fig:digest}
\end{figure}

Although interaction controllers have long been studied in classical robotics, learning their parameters jointly with robot motion has received comparatively limited attention in learning-based manipulation. In particular, impedance control provides a well-established approach for regulating a robot’s response to physical interaction by defining the relationship between motion errors and interaction forces through stiffness and damping parameters. Varying stiffness\footnote{Because damping is designed based on the selected stiffness in our implementation, we refer primarily to stiffness throughout this paper. Nevertheless, the choice of damping is important in contact-rich manipulation, as it strongly influences transient behavior, stability, and contact-force regulation.} throughout a task enables the robot to trade off accurate motion tracking and compliant interaction according to the current task requirements. Incorporating such behavior into imitation learning, however, requires suitable stiffness targets to be associated with the demonstrated motion.

Existing approaches obtain stiffness targets either through manual selection during demonstration \cite{COMP-ACT, DIPCOM} or by deriving compliance after the demonstrations have been collected \cite{11128452}. The former increases the burden on the operator, while the latter can introduce a mismatch between data collection and deployment: applying a different compliance profile changes the robot's dynamics and can therefore shift the state distribution encountered by the policy. In imitation learning, such distribution shifts are particularly problematic \cite{NEURIPS2021_07d59386} because deviations from the demonstrated trajectories can compound during execution. This motivates generating and applying stiffness during the demonstration, so that motion and compliance are recorded under the same interaction dynamics that the policy is expected to reproduce.

In this work, we propose Imp-ACT, an ACT-based imitation learning framework that combines human-demonstrated motion with direction-dependent stiffness generated online by an adaptive Cartesian impedance controller (see Fig.~\ref{fig:digest}). During data collection, the human operator commands only the desired end-effector pose and gripper action through a virtual reality interface, while the controller automatically regulates translational stiffness based on interaction feedback using an adaptation law based on~\cite{pietro}. The resulting stiffness commands are applied to the robot and recorded synchronously with the demonstrated motion. This preserves an intuitive, kinematics-based demonstration interface while ensuring that the resulting robot behavior is compliant and adaptive by design.

From these demonstrations, Imp-ACT learns to jointly predict end-effector pose, gripper action, and motion-direction stiffness from visual observations, robot state, and interaction wrench. During autonomous execution, the learned stiffness prediction replaces the demonstration-time adaptation mechanism, allowing motion and compliance to be generated jointly by the policy.


\section{Related Work}

\subsection{Variable Impedance Control}

Impedance control is widely used in contact-rich tasks such as assembly, rehabilitation, and human-robot co-manipulation \cite{loris2019assembly,jamwal2016rehab,ficuciello2014manip}. Conventional impedance controllers typically use fixed stiffness and damping parameters \cite{song2019review}, limiting their ability to adapt to changing task conditions. Variable impedance control addresses this by adjusting these parameters online according to task and interaction requirements \cite{abudakka2020variable}.

In the literature, several strategies have been proposed for adapting impedance parameters. Some approaches focus on task-specific adaptation, including a fuzzy-neural-network controller for upper-limb rehabilitation~\cite{xu2011adaptive} and gait-phase-based impedance regulation for a powered prosthetic knee~\cite{mohammadi2019variable}. Alternatively, teleimpedance methods transfer stiffness estimated from the operator's EMG activity~\cite{ajoudani2012teleimpedance,yang2015teleimpedance, ozdamar2022shared}. While these methods can be tailored to a variety of tasks, EMG-based estimation requires additional sensors and calibration, limiting its practicality. A more generic approach is introduced by Balatti et al.~\cite{pietro}, where a self-tuning Cartesian impedance controller is proposed to adapt its stiffness and damping online based on interaction forces and tracking errors, while regulating the impedance parameters only along the direction of motion. The scalability of this approach was tested in different scenarios, including agricultural scooping, pallet-jack manipulation~\cite{pietro}, and debris removal~\cite{pietrodebris}.

\subsection{Imitation Learning}
Imitation learning enables robots to acquire manipulation skills from demonstrations rather than explicit programming \cite{lfd_review}. Modern visuomotor policies such as ACT \cite{ACT} and Diffusion Policy \cite{chi2024diffusionpolicy} map visual and proprioceptive observations to action sequences, while larger models such as OpenVLA \cite{pmlr-v270-kim25c} and $\pi_{0.5}$ \cite{pmlr-v305-black25a} target broader generalization. Despite these advances, such policies primarily predict motion commands, leaving the robot's response to contact to the low-level controller rather than learning it as part of the behavior.

A first line of work incorporates interaction feedback into the policy observations. CRAFT \cite{zhang2026craft} introduces a force-aware curriculum that encourages the policy to exploit joint-torque feedback, while TacVLA \cite{zhang2026tacvla} integrates tactile information only when contact is detected. These approaches improve contact awareness, but force or tactile information only influences the predicted motion; the compliance with which that motion is executed remains determined by the low-level controller.

Other approaches incorporate compliance into the action space. Comp-ACT \cite{COMP-ACT} and DIPCOM \cite{DIPCOM} jointly predict motion and stiffness but require manual selection of predefined compliance modes during teleoperation. Adaptive Compliance Policy (ACP) \cite{11128452} removes this burden by reconstructing compliance targets offline from forces measured during uniformly low-stiffness demonstrations. CompliantVLA-adaptor \cite{zhang2026compliantvla} instead introduces compliance at deployment, pairing a pretrained VLA with a separate module for stiffness and damping selection.

Although reconstructed equilibrium targets can reproduce the reference force at demonstrated poses, changes in compliance can alter the robot’s response to tracking errors and contact disturbances. Introducing compliance after collection can therefore create a mismatch between demonstration and execution dynamics, potentially shifting the states encountered by the policy. Such distribution shifts can contribute to compounding execution errors \cite{NEURIPS2021_07d59386}. This motivates generating and applying stiffness during teleoperation, so that motion and compliance targets are recorded within the same closed-loop interaction.

Imp-ACT implements this collection strategy using a self-tuning Cartesian impedance controller \cite{pietro}. While the operator commands only the end-effector pose and gripper action, the controller adapts stiffness along the motion direction while maintaining compliance in the orthogonal directions. The applied stiffness is recorded alongside the commanded motion, without manual stiffness selection or offline reconstruction. Imp-ACT then learns to jointly predict motion, gripper action, and stiffness from visual, proprioceptive, and wrench observations. This preserves the correspondence between the recorded stiffness targets and the controller settings used during collection.


\begin{figure*}[!t]
\vspace*{2mm}
\centering
\includegraphics[width=1\textwidth]{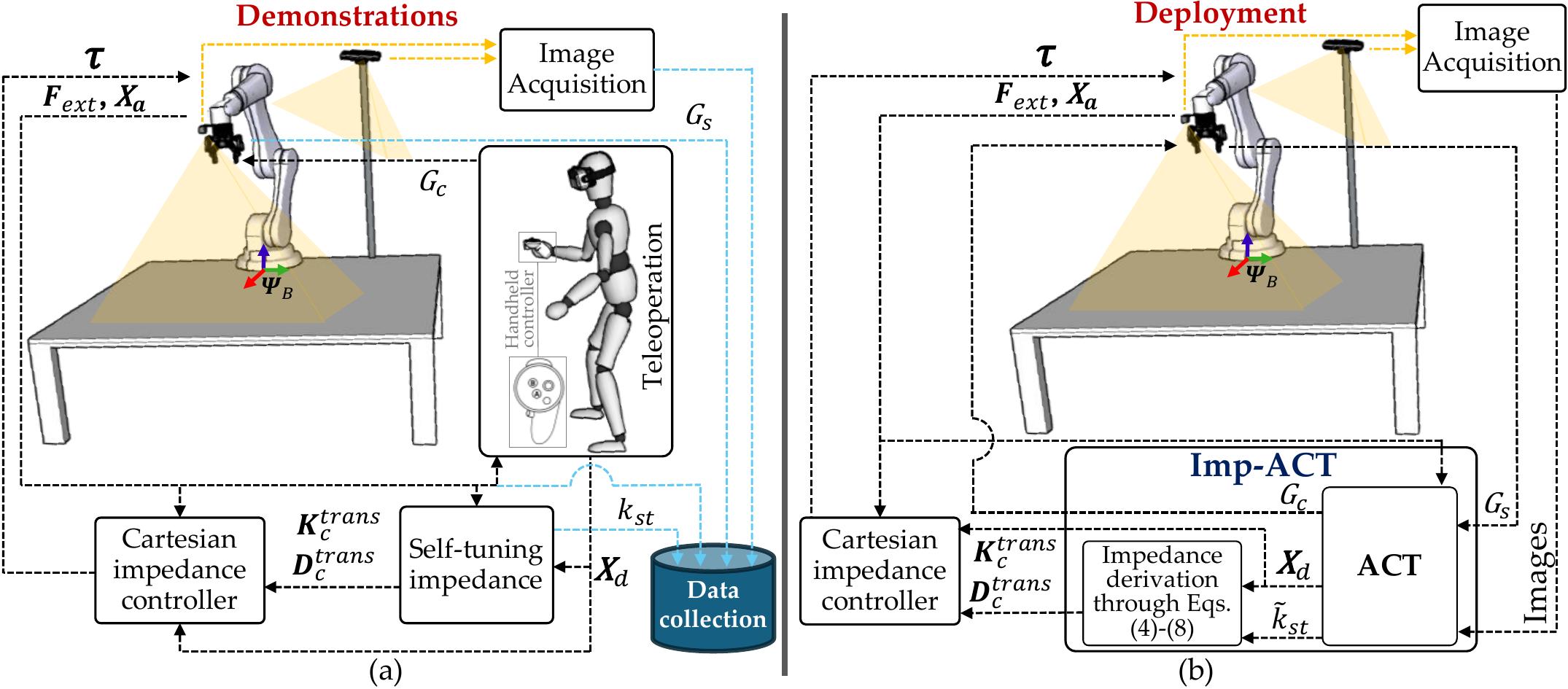}
\caption{Overview of (a) Demonstrations and (b) Deployment phases, where $\bm{\Psi}_B$ denotes the base frame. During demonstration collection, images from the cameras, measured end-effector pose $\boldsymbol{X}_a$, external interaction wrench $\boldsymbol{F}_{ext}$, and the gripper state $G_s$  are recorded as observations. The corresponding action targets comprise the desired end-effector pose $\boldsymbol{X}_d$, the gripper command $G_c={open, close}$, and the applied stiffness coefficient $k_{st}$. During deployment, Imp-ACT predicts these action targets, including the stiffness coefficient $\tilde{k}_{st}$.
}

\label{fig:flow}
\end{figure*}

\section{Methodology}

The proposed framework enables the policy to jointly learn desired end-effector poses and task-appropriate stiffness behavior from human demonstrations. The methodology is presented in four parts: \textit{Cartesian Impedance Controller}, \textit{Self-Tuning Impedance Approach}, \textit{Demonstration Collection and Representation}, and \textit{Imp-ACT Policy Learning and Deployment}. The first two describe how compliant robot behavior is generated and adapted online during task execution, while the latter two describe how demonstrations are processed and used to train and deploy the Imp-ACT policy (see Fig.~\ref{fig:flow}).

\subsection{Cartesian Impedance Controller}

Cartesian impedance control allows the robot to modulate its end-effector dynamics in response to external forces,
enabling a compliant and controlled interaction ~\cite{schindlbeck2015unified, ajoudani2017choosing}. This approach relies on real-time torque sensing and actuation, with the robot rigid body model in joint space described by:
\begin{equation}
  \boldsymbol{\tau} =  \boldsymbol{M(q)\ddot{q}} + \boldsymbol{C(q,\dot{q})\dot{q}} + \boldsymbol{g(q)}   + \boldsymbol{\tau}_{ext},
\end{equation}
\begin{equation}
  \boldsymbol{\tau}_{ext} =  \boldsymbol{J(q)}^T \boldsymbol{F}_c + \boldsymbol{\tau}_{st},
\end{equation}
where $n$ is the number of joints, $\boldsymbol{q} \in \mathbb{R}^{n}$ is the joint-angle vector, $\boldsymbol{J} \in \mathbb{R}^{6 \times n}$ is the robot arm Jacobian matrix, $\boldsymbol{M} \in \mathbb{R}^{n \times n}$ is the mass matrix, $\boldsymbol{C} \in \mathbb{R}^{n \times n}$ is the Coriolis and centrifugal matrix, $\boldsymbol{g} \in \mathbb{R}^{n}$ is the gravity vector and $\boldsymbol{\tau}_{ext}$ is the external torque vector. $\boldsymbol{F}_c$ represents the wrench vector in the Cartesian space and $\boldsymbol{\tau}_{st}$ the second task torques projected onto the null-space of $\boldsymbol{J}$. 
The Cartesian wrench $\boldsymbol{F}_c \in \mathbb{R}^{6}$ is calculated as follows:
\begin{equation}
  \boldsymbol{F}_c =  \boldsymbol{K}_c (\boldsymbol{X}_d - \boldsymbol{X}_a) + \boldsymbol{D}_c (\boldsymbol{\dot{X}}_d - \boldsymbol{\dot{X}}_a),
\end{equation}
\noindent where $\boldsymbol{K}_c \in \mathbb{R}^{6 \times 6}$ and $\boldsymbol{D}_c \in \mathbb{R}^{6 \times 6}$ represent the Cartesian stiffness and damping matrices, respectively. $\boldsymbol{X}_d$ and $\boldsymbol{X}_a \in \mathbb{R}^{6}$ are the desired and actual Cartesian poses, $\boldsymbol{\dot{X}}_d$ and $\boldsymbol{\dot{X}}_a \in \mathbb{R}^{6}$ denote their corresponding velocity profiles.

\subsection{Self-Tuning Impedance Approach}

For the online stiffness calculation, we adopt the core principles of the self-tuning impedance controller presented in \cite{pietro}, which assigns higher stiffness along the direction of motion, enabling the robot to track the commanded trajectory and reach the desired goal position, while maintaining lower stiffness along the orthogonal directions. This allows the robot to adjust its motion in response to contact forces, environmental constraints, and unintended disturbances. To this end, the Cartesian stiffness matrix $\boldsymbol{K}_c$ and the damping matrix $\boldsymbol{D}_c$ are tuned online according to the desired translational motion direction, defined as:
\begin{equation}\label{p_vector}
  \overrightarrow{\boldsymbol{P}}(t) =  \boldsymbol{X}_{d}^{trans}(t) - \boldsymbol{X}_{d}^{trans}(t-1).
\end{equation}
Along the two axes orthogonal to $\overrightarrow{\boldsymbol{P}} \in \mathbb{R}^3$, compliant behavior is maintained by setting the corresponding stiffness and damping coefficients to ${k}_{min}$ and $d_{min} = 2\zeta\sqrt{k_{min}}$, respectively \cite{albu2003cartesian}.
To implement this direction-dependent adaptive impedance behavior, the symmetric and positive-definite translational components of the Cartesian stiffness and damping matrices, $\boldsymbol{K}_{c}^{trans} \in \mathbb{R}^{3 \times 3}$ and $\boldsymbol{D}_{c}^{trans} \in \mathbb{R}^{3 \times 3}$, are derived as follows: 

\begin{equation}\label{eq:modalStiffness}
    \boldsymbol{K}_{c}^{trans} = \boldsymbol{U}\boldsymbol{\Sigma}_{k}\boldsymbol{U}^{T}, 
\end{equation}
\begin{equation}\label{eq:modalDamping}
    \boldsymbol{D}_{c}^{trans} = \boldsymbol{U}\boldsymbol{\Sigma}_{d}\boldsymbol{U}^{T},
\end{equation}
where $\boldsymbol{\Sigma}_{k} \in \mathbb{R}^{3 \times 3}$ and $\boldsymbol{\Sigma}_{d} \in \mathbb{R}^{3 \times 3}$ are diagonal matrices containing the desired stiffness and damping parameters along the directions of the vectors composing the orthonormal basis $\boldsymbol{U} \in \mathbb{R}^{3 \times 3}$. The first column of $\boldsymbol{U}$ corresponds to the normalized desired motion vector, $\hat{\boldsymbol{P}} = \overrightarrow{\boldsymbol{P}} / \lVert\overrightarrow{\boldsymbol{P}}\rVert$, while the remaining two columns are derived such that they form an orthonormal basis. $\boldsymbol{\Sigma}_{k}$ and $\boldsymbol{\Sigma}_{d}$ are defined as:

\begin{equation}\label{eq:sigma-k}
    \boldsymbol{\Sigma}_{k} = \textrm{diag}({k_{st}},{k_{min}},{k_{min}}),
\end{equation}
\begin{equation}\label{eq:sigma-d}
    \boldsymbol{\Sigma}_{d} = \textrm{diag}({d_{st}},{d_{min}},{d_{min}}),
\end{equation}
where $k_{st} \in \mathbb{R}_{>0}$ is the self-tuning stiffness coefficient applied along the motion direction $\overrightarrow{\boldsymbol{P}}$ and bounded by $k_{\min}\leq k_{st}\leq k_{\max}$. The corresponding damping coefficient ${d_{st}}$ is  calculated as $d_{st} = 2\zeta\sqrt{k_{st}}$ \cite{albu2003cartesian}.  

At each time step $t$, $k_{st}$ is updated as follows:

\begin{equation}
        k_{st}(t) = \min\Bigl\{ k_{max}, \; k_{st}(t-1) + \alpha\|\Delta\boldsymbol{X}_{err}(t)\|\Delta T \Bigr\}, \label{eq:kst_increase}
\end{equation}
\begin{equation}
    \Delta\boldsymbol{X}_{err}(t) = \boldsymbol{X}_d^{trans}(t) - \boldsymbol{X}^{trans}_a(t),
\end{equation}
where $\alpha$ is the update parameter, $\|\Delta\boldsymbol{X}_{err}\|$ is the Euclidean norm of the end-effector translational tracking error, and $\Delta T$ is the control-loop sampling period. It is important to note that, $k_{st}$ is subject
to changes only when the $\|\Delta\boldsymbol{X}_{err}\|$ exceeds the position error threshold $\Delta{X}^{th}_{err}\in \mathbb{R}_{> 0}$.
This threshold introduces an error tolerance that prevents unnecessary increases in impedance and stops the stiffness growth once the desired tracking accuracy is reached.

On the other hand, maintaining high stiffness is mostly not desired when the robot leaves the contact region and continues its motion in free space. This situation is characterized by a small tracking error accompanied by a decrease in the external interaction force. To this end, the translational force drop rate $\dot F_{ext} \in \mathbb{R}$ is computed as:
\begin{equation}\dot{F}_{ext}(t) = \left(\|\boldsymbol{F}_{ext}^{trans}(t-1)\| - \|\boldsymbol{F}_{ext}^{trans}(t)\|\right)/{\Delta T},
\end{equation} 
where $\|\boldsymbol{F}_{ext}^{trans}\|\in \mathbb{R} $ is the magnitude of the external force at the end-effector. Note that positive $\dot F_{ext}$ indicates a decrease in the interaction force, while a negative value indicates the opposite. Whenever the Cartesian position error remains below its threshold  ($\|\Delta\boldsymbol{X}_{err}\|<\Delta X^{th}_{err}$) and the force drop rate exceeds its threshold $\dot F_{ext} > \dot F^{th}_{ext}$ the stiffness is decreased according to:
\begin{equation}
    k_{st}(t) =\max\Bigl\{k_{\min}, k_{st}(t-1) - \alpha10^{-2}\dot F_{ext}(t)\Delta T \Bigr\}.
    \label{eq:kst_decrease}
\end{equation}
This prevents the stiffness from remaining unnecessarily high once the external interaction force has diminished. Here, $\alpha$ is scaled by a factor of $10^{-2}$, to account for the different scales of the position-error and force-drop-rate signals. The pseudocode in Algorithm \ref{alg:alg_1} summarizes the complete flow of the implemented self-tuning impedance approach.


\begin{algorithm}[!h]
\caption{Self-tuning impedance algorithm}\label{alg:alg_1}
\begin{algorithmic}
\setstretch{1.2}

\Require $\overrightarrow{\boldsymbol{P}}$, $\Delta\boldsymbol{X}_{err}$, $\dot F_{ext}$
\Ensure $\boldsymbol{K}_c^{trans}$ , $\boldsymbol{D}_{c}^{trans}$

\State \emph{Initialization:}
\State 
$k_{st}(0)=k_{min}$,  $d_{st}(0)=d_{min}$, \\
$\boldsymbol{\Sigma}_{k}= k_{min}\boldsymbol{I_{3\times3}}$ , $\boldsymbol{\Sigma}_{d}= d_{min}\boldsymbol{I_{3\times3}}$
\State \emph{Control loop:}

\If{$\|\Delta\boldsymbol{X}_{err}(t)\| > \Delta X_{err}^{th}$} \Comment{Increase stiffness}
    \State $k_{st}(t) = \min\left\{ k_{max}, \; k_{st}(t-1) + \alpha\|\Delta\boldsymbol{X}_{err}(t)\|\Delta T \right\}$
\ElsIf {$\dot{F}_{ext}(t) > \dot{F}_{ext}^{th}$} \Comment{Decrease stiffness}
    \State $k_{st}(t) =\max\left\{k_{\min}, k_{st}(t-1) - \alpha10^{-2}\dot F_{ext}(t)\Delta T \right\}$ 
\EndIf
\State $\boldsymbol{\Sigma}_{k,1,1}= k_{st}(t)$, $\boldsymbol{\Sigma}_{d,1,1}= 2\zeta\sqrt{k_{st}(t)}$
\State $\boldsymbol{U}=\textit{getOrthonormalBasis}(\overrightarrow{\boldsymbol{P}})$
\State $\boldsymbol{K}_{c}^{trans}= \boldsymbol{U}\boldsymbol{\Sigma}_{k}\boldsymbol{U}^{T}$, $\boldsymbol{D}_{c}^{trans}= \boldsymbol{U}\boldsymbol{\Sigma}_{d}\boldsymbol{U}^{T}$  \Comment{Output}

\end{algorithmic}
\end{algorithm}
\setlength{\textfloatsep}{8pt plus 2pt minus 3pt}

\subsection{Demonstration Collection and Representation}
Demonstrations are collected through teleoperation and stored using the LeRobot~\cite{cadene2024lerobot} dataset format, in which each recorded frame pairs the robot and sensor observations with the corresponding action target. At each recording step, the observations comprise the measured tool-center-point (TCP) pose, the gripper state (motor current as a proxy for grasping force and a Boolean object-detection flag), the latest available camera images, and the external wrench at the TCP that is obtained by mapping the measurements from the force/torque (F/T) sensor mounted between the robot flange and the gripper.
The corresponding action target contains the processed Cartesian equilibrium pose and gripper command. During demonstrations, the stiffness value \(k_{st}\), obtained from the controller's online adaptation, is additionally appended to the action vector without requiring explicit input from the operator. Following Zhou et al. \cite{Zhou_2019_CVPR}, measured and target orientations use a continuous 6D representation comprising the first two rotation-matrix rows, avoiding rotation-vector discontinuities at \(\pi\). At inference, rotation predictions are decoded via Gram–Schmidt orthonormalization and converted to the low-level controller’s pose representation.
\subsection{Imp-ACT Policy Learning and Deployment}

Imp-ACT extends ACT’s action space with $\tilde{k}_{st}$ and predicts 50-step action chunks, corresponding to a one-second horizon. The observation space comprises two \(480\times640\) RGB images from a wrist-mounted camera and a fixed camera behind the robot, plus an 18-dimensional state: end-effector position (3D), orientation (6D), gripper position, force/torque wrench (6D), gripper motor current (a proxy for grasping force), and a binary object-detection flag.

Each 11-dimensional action contains the Cartesian equilibrium position (3D), orientation (6D), gripper command, and motion-direction stiffness parameter $\tilde{k}_{st}$. Stiffness in the two orthogonal translational directions remains fixed. The policy runs at \(50\,\mathrm{Hz}\), using temporal ensembling with a weighting coefficient of \(0.1\) to smooth overlapping predictions. Each action dimension, including stiffness, is normalized using its mean and standard deviation computed from the training demonstrations. At deployment, predicted actions are converted back to their original units before being passed to the low-level controller.
Eqs.~\eqref{p_vector}--\eqref{eq:sigma-d} map $\tilde{k}_{st}$ to the translational Cartesian stiffness and damping used by the low-level controller.


\section{Experiments}
\subsection{Experimental Setup}

Experiments are conducted using a torque-controlled 7-DoF Franka Emika Panda robotic arm equipped with a Robotiq 2F-85 two-finger parallel gripper. To enable precise measurement of external contact forces, an additional six-axis ATI Mini-45 F/T sensor is mounted between the robot's flange and the gripper. 
Visual observations are provided by a wrist camera (Intel RealSense D435i) attached adjacent to the gripper for capturing close-range contact interactions, and a back camera (Logitech B525 HD) positioned with an offset of (-5, 41, 56) cm along the X-, Y-, and Z-axes of the robot base frame for obtaining broader task context.
Demonstrations are collected via teleoperation, using a Meta Quest 3 headset as the interface (see Fig.~\ref{fig:digest} for an overview of the setup). The system is integrated within the Robot Operating System 2 (ROS 2), where RGB camera streams are acquired at 30 fps and teleoperation commands are updated at 72 Hz, while low-level robot control is performed through the Franka Control Interface (FCI), which communicates with the robot over Ethernet at 1 kHz. Policy training was performed on an HPC cluster. Each training job was allocated 20 CPU cores and 64~GB of RAM. The nodes provide four NVIDIA Tesla V100 GPUs with 16~GB of memory each. Inference was run on a Lenovo Legion 9i Gen 10 laptop equipped with an Intel Core Ultra 9 275HX processor and an NVIDIA GeForce RTX 5090 Laptop GPU.

\subsection{Teleoperation Interface}
Teleoperation is conducted using a custom Unity passthrough application on a VR headset, connected to a recording framework based on ROS 2 and LeRobot. The application streams six-degree-of-freedom poses and button states from a four-button handheld controller, mapping motion into robot coordinates using the session-verified transformation of Rosasco et al. \cite{pmlr-v305-rosasco25a, 10000218} (see Fig. \ref{fig:flow}).

A grip button acts as a clutch for relative control of the Cartesian equilibrium target. On engagement, the current handheld controller and end-effector poses are stored as references; subsequent controller translations and rotations update the target relative to these poses. Releasing the clutch pauses target updates and recording while retaining the last target, allowing hand repositioning. The analog trigger proportionally controls gripper aperture, while two buttons terminate and retain or discard the demonstration.

A wrist-camera panel, toggled via the thumbstick, provides a view of otherwise occluded contact regions. An augmented-reality arrow displays the interaction force, while a gauge indicates whether its magnitude lies below, within, or above a predefined band. These cues help the operator unload contact and retract the equilibrium target before releasing an inserted object, reducing the risk of unintended end-effector motion into the surrounding structure.

\subsection{Experimental Scenarios}
To evaluate the effectiveness of Imp-ACT, we compare it against two fixed-stiffness baselines: a low-stiffness compliant controller and a high-stiffness controller. The compliant baseline uses a fixed translational stiffness of $200$ N/m, whereas the high-stiffness baseline uses $1000$ N/m.
We evaluate the proposed approach against the baselines on two contact-rich manipulation scenarios, namely 1) \textit{wiping} and 2) \textit{plug insertion}, each imposing distinct interaction and compliance requirements. For each task, each policy was trained on a separate demonstration dataset collected using its corresponding controller. The number of demonstrations, task-variation protocol, training settings, and evaluation conditions were identical across policies. The operator aimed to demonstrate each task consistently across conditions, without deliberately adjusting their strategy to favor any controller.

\textit{1) Wiping:} The goal of this task is to remove marks from a whiteboard, with their positions and shapes varying across episodes. Each episode begins with the robot at its home configuration. The teleoperator guides the robot to grasp the eraser, wipe the board until no visible trace of the mark remains, and return the eraser to its initial location (see Fig.~\ref{fig:ellipsoid-tasks}(a)-(d)). We collected 40 demonstration episodes under each of three stiffness conditions: fixed low stiffness, fixed high stiffness, and adaptive stiffness using the proposed approach. Stiffness regulation is important for maintaining effective wiping and consistent erasing, while limiting excessive contact forces at high $K_c$ and stick-slip-induced vibration in both baseline conditions. This scenario therefore provides a basis for comparing the three conditions in terms of their ability to maintain the surface contact required for effective wiping as the locations and shapes of the marks vary across episodes.

\textit{2) Plug insertion:} The goal of this task is to insert a two-pin plug into the central socket of a five-outlet power strip. As in the wiping task, each episode begins with the robot at its home configuration. The teleoperator then guides the robot to grasp the plug, move it toward the power strip, align the plug with the central socket, and complete the insertion (see Fig.~\ref{fig:ellipsoid-tasks}(e)-(h)). To introduce variability across demonstrations, the initial position of the plug was randomized within $\pm2\,\text{cm}$ in translation and $\pm10^\circ$ in rotation about the X-axis of the robot base frame. Because plug insertion requires greater precision than wiping, 80 episodes were collected under each stiffness condition (low, high, and adaptive) for this task. Note that the need to simultaneously align both pins of the plug with the two holes of the socket, $\approx 1\,\text{mm}$ tolerance, poses a greater challenge than classic peg-in-hole insertion tasks widely used in the literature to represent complex interactions with the environment. 
Consequently, even minor misalignments at contact, whether due to tracking inaccuracies, sensor noise, or similar factors, may generate large interaction forces or prevent successful insertion, highlighting the need to balance positional accuracy with compliant behavior. Video of these experiments can be found at: 
\url{https://youtu.be/iAu_HFeaCRg}

\begin{figure*}[t]
  \vspace*{2mm}
  \centering

  \begin{subfigure}[t]{0.21\textwidth}
    \begin{overpic}[width=\linewidth,height=\panelheight]
      {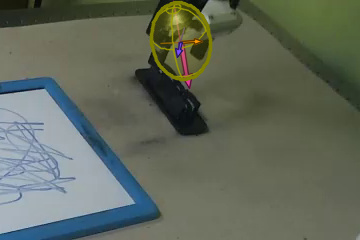}
      \timelabel{$t=1.4$\,s}
      \kstlabel{$\tilde{k}_{st}=300$\,N/m}
    \end{overpic}
    \caption{Free.}
    \label{fig:wipe-a}
  \end{subfigure}\hfill
  \begin{subfigure}[t]{0.21\textwidth}
    \begin{overpic}[width=\linewidth,height=\panelheight]
      {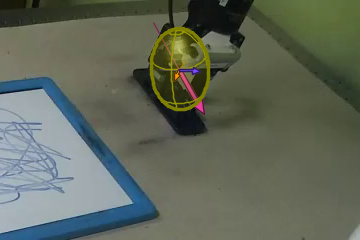}
      \timelabel{$t=3.4$\,s}
      \kstlabel{$\tilde{k}_{st}=327$\,N/m}
    \end{overpic}
    \caption{Descending.}
    \label{fig:wipe-b}
  \end{subfigure}\hfill
  \begin{subfigure}[t]{0.21\textwidth}
    \begin{overpic}[width=\linewidth,height=\panelheight]
      {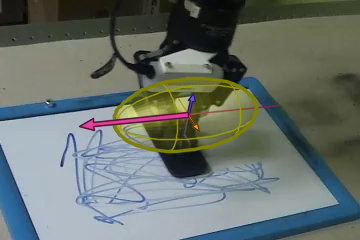}
      \timelabel{$t=9.1$\,s}
      \kstlabel{$\tilde{k}_{st}=464$\,N/m}
    \end{overpic}
    \caption{Wiping.}
    \label{fig:wipe-c}
  \end{subfigure}\hfill
  \begin{subfigure}[t]{0.21\textwidth}
    \begin{overpic}[width=\linewidth,height=\panelheight]
      {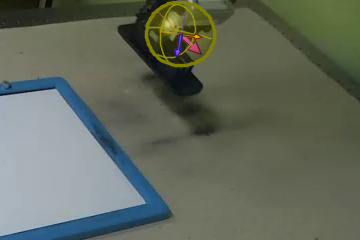}
      \timelabel{$t=24.4$\,s}
      \kstlabel{$\tilde{k}_{st}=360$\,N/m}
    \end{overpic}
    \caption{Wiping complete.}
    \label{fig:wipe-d}
  \end{subfigure}

  \par\vspace{0.35em}

  \begin{subfigure}[t]{0.21\textwidth}
    \begin{overpic}[width=\linewidth,height=\panelheight]
      {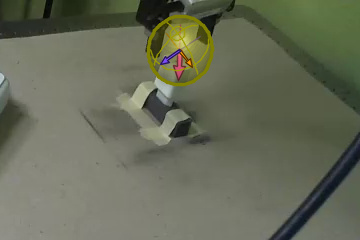}
      \timelabel{$t=2.6$\,s}
      \kstlabel{$\tilde{k}_{st}=237$\,N/m}
    \end{overpic}
    \caption{Over the plug.}
    \label{fig:plug-a}
  \end{subfigure}\hfill
  \begin{subfigure}[t]{0.21\textwidth}
    \begin{overpic}[width=\linewidth,height=\panelheight]
      {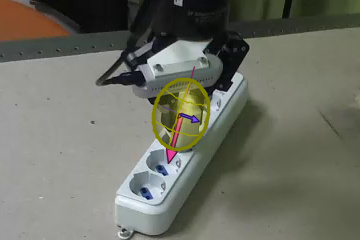}
      \timelabel{$t=12.8$\,s}
      \kstlabel{$\tilde{k}_{st}=277$\,N/m}
    \end{overpic}
    \caption{Inserting.}
    \label{fig:plug-b}
  \end{subfigure}\hfill
  \begin{subfigure}[t]{0.21\textwidth}
    \begin{overpic}[width=\linewidth,height=\panelheight]
      {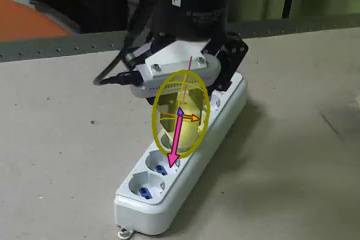}
      \timelabel{$t=14.0$\,s}
      \kstlabel{$\tilde{k}_{st}=321$\,N/m}
    \end{overpic}
    \caption{Fully inserted.}
    \label{fig:plug-c}
  \end{subfigure}\hfill
  \begin{subfigure}[t]{0.21\textwidth}
    \begin{overpic}[width=\linewidth,height=\panelheight]
      {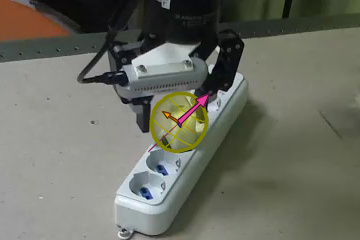}
      \timelabel{$t=18.4$\,s}
      \kstlabel{$\tilde{k}_{st}=208$\,N/m}
    \end{overpic}
    \caption{Released.}
    \label{fig:plug-d}
  \end{subfigure}

  \caption{Cartesian stiffness ellipsoids during wiping (top) and plug insertion (bottom). The proposed approach increases the predicted stiffness $\tilde{k}_{st}$ along the instantaneous motion direction (magenta arrow), while maintaining the two orthogonal directions at $200$\,N/m. The principal stiffness direction follows the task motion, lying approximately along the wiping direction or insertion axis. Each panel reports the time (upper left) and $\tilde{k}_{st}$ (lower right). Across the illustrated episodes, $\tilde{k}_{st}$ ranges from $200$--$642$\,N/m for wiping and from $200$--$329$\,N/m for insertion.}
  \label{fig:ellipsoid-tasks}
  
\end{figure*}

\subsection{System Parameters}

In the adaptive-stiffness case, $k_{st}$, representing the stiffness along the motion direction $\overrightarrow{\boldsymbol{P}}$, was bounded within $k_{\min}=200$ N/m and $k_{\max}=1000$ N/m. Along the other two directions orthogonal to $\overrightarrow{\boldsymbol{P}}$, the stiffness was fixed at $k_{\min}=200$ N/m, thereby preserving compliance in the plane perpendicular to the primary motion direction and allowing the end-effector to accommodate small deviations without unnecessary resistance. 
For the rotational components, stiffness was fixed at $30$ Nm/rad along all three axes. 
The Cartesian stiffness matrices of the low and high stiffness baselines were deliberately set to $\boldsymbol{K}_c^{{low}}=\operatorname{diag}(200,200,200,30,30,30)$ and $\boldsymbol{K}_c^{{high}}=\operatorname{diag}(1000,1000,1000,30,30,30)$, respectively, to match the lower and upper bounds of the adaptive stiffness range, enabling a fair comparison across all cases. Following this, the damping was set according to $\boldsymbol{D}_c=2\zeta \sqrt{\boldsymbol{K}_c}$ with $\zeta=0.7$, for all conditions. 
The stiffness adaptation gain was selected as $\alpha=1000$ to balance rapid adaptation to external interactions against abrupt stiffness changes.
The Cartesian position error and force-drop-rate thresholds were empirically chosen as $\Delta X^{th}_{err}=5$ cm and $\dot F^{th}_{ext}=2$ N/s, respectively. These values were not highly sensitive to small variations, as their role is to detect meaningful changes in the interaction. For both tasks, each policy was trained for 300,000 steps, taking approximately 16 hours per run.


\section{Results}
\label{sec:results}
We compare the three stiffness strategies in terms of task success and interaction performance, evaluating each policy over 25 trials per task. Interaction metrics are computed only over successful episodes (\(24\), \(25\), and \(25\) for wiping, and \(15\), \(18\), and \(19\) for plug insertion) to ensure a consistent comparison across the three policies.

\subsection{Stiffness adaptation}
To illustrate how the learned stiffness changes during task execution, Fig.~\ref{fig:ellipsoid-tasks} visualizes the corresponding Cartesian stiffness ellipsoid for representative wiping and plug-insertion episodes. The top row reports the wiping sequence, while the bottom row shows the approach, insertion, and release phases of the plug task. In both cases, the main axis of the ellipsoid is aligned with the instantaneous direction of desired end-effector motion, indicated by the magenta arrow, while the two orthogonal directions remain at the compliant baseline. Higher interaction forces are associated with larger values of $\tilde{k}_{st}$, making the robot stiffer along the direction of motion. This stiffening is more pronounced during wiping, which involves sustained contact, with respect to plug insertion. 

\begin{figure}[t]
  \centering\includegraphics[width=1\linewidth]{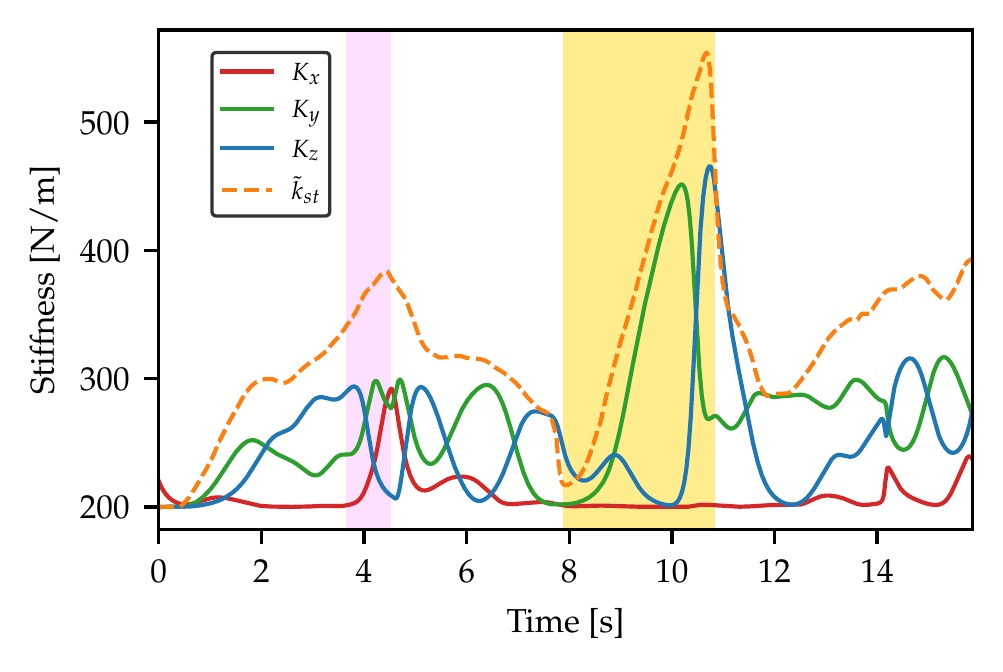}
  
  \caption{Predicted Cartesian stiffness during one wiping episode. $K_x$, $K_y$ and $K_z$ are the stiffness commanded w.r.t $\boldsymbol{\Psi}_B$, and $\tilde{k}_{st}$ (dashed) the inferred stiffness along
  the direction of motion. The pink region marks the gripper closing on the wiper while the yellow one marks contact with the board.}
  
  \label{fig:stiffness-wiping}
\end{figure}

To further examine the stiffness evolution during task execution, Fig.~\ref{fig:stiffness-wiping} reports \(\tilde{k}_{st}\) together with the Cartesian stiffness components \(K_x\), \(K_y\), and \(K_z\) for a representative wiping episode. The policy continuously modulates the Cartesian stiffness along the instantaneous direction of motion through \(\tilde{k}_{st}\), while the stiffness in the two orthogonal directions remains fixed at $200$ N/m. Stiffness adaptation remains active throughout the episode, rather than only during contact. During free-space motion and grasping, $\tilde{k}_{st}$ averages $317$~N/m, improving Cartesian tracking relative to the compliant baseline. Once contact with the surface is established, it increases to an average of $377$ N/m and peaks at $554$ N/m before decreasing as contact is released. The policy therefore continuously regulates the tracking--compliance trade-off rather than switching between discrete free-space and contact regimes.

The amount of adaptation also reflects the task requirements. Across successful wiping episodes, $\tilde{k}_{st}$ peaks at $547$ N/m on average (range $485$--$705$ N/m), compared with $299$ N/m for plug insertion (range $272$--$329$ N/m). Wiping requires sustained surface interaction, whereas plug insertion involves shorter and more localized contacts.

\begin{figure*}[t]
    \vspace*{2mm}
    \centering
    \includegraphics[width=1\textwidth]{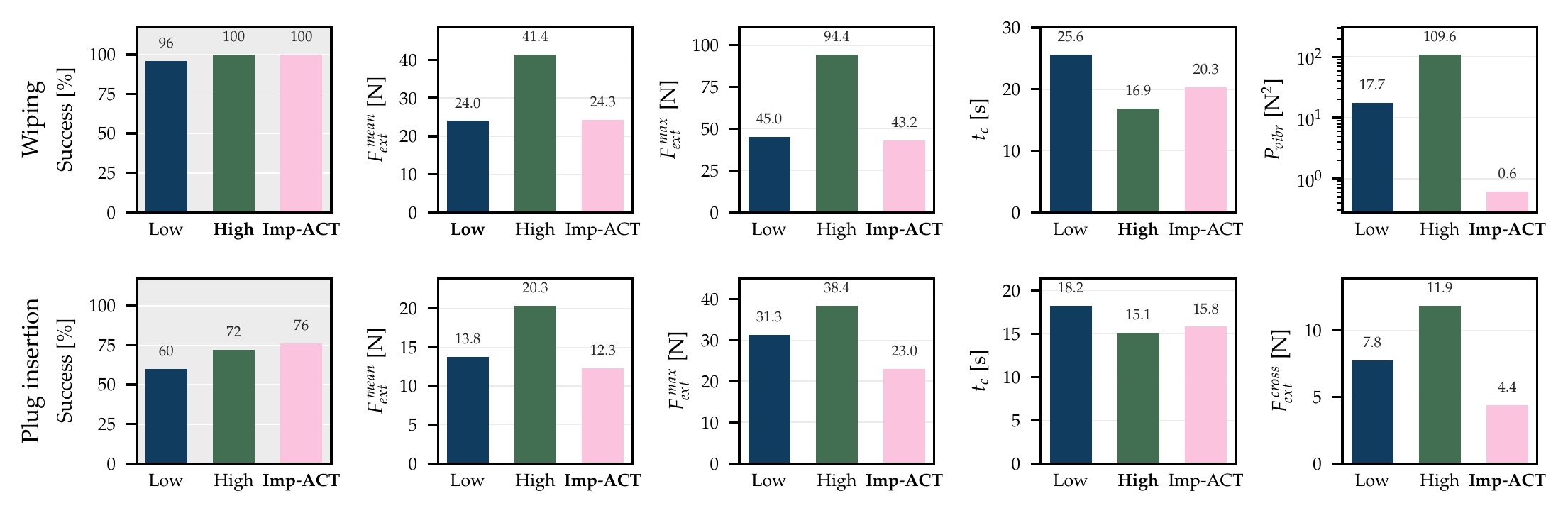}
    \caption{Task outcome and contact metrics for wiping (top) and plug insertion (bottom). Grey-background metrics are better when higher; white-background metrics are better when lower and are computed over successful episodes only. Bold text indicates the best result, including ties.
    \label{fig:contact-metrics}
}

\end{figure*}
\subsection{Wiping}

In the wiping task, both the high-stiffness baseline and Imp-ACT achieve a \(100\%\) success rate over \(25\) trials (see Fig.~\ref{fig:contact-metrics}), while the low-stiffness baseline achieves \(96\%\), failing once during grasp acquisition. This failure may be attributed to the poorer pose-tracking accuracy associated with uniformly low Cartesian stiffness.

The main differences emerge in interaction behavior. In Fig.~\ref{fig:contact-metrics}, \(F_{{ext}}^{{mean}}\) denotes the average of the mean interaction forces computed for individual episodes, while \(F_{{ext}}^{{max}}\) is the corresponding average of the per-episode peak interaction forces. The low-stiffness baseline yields the smallest $F_{{ext}}^{{mean}}$, which is $24.0$~N, closely followed by Imp-ACT with $24.3$~N, while the high-stiffness baseline reaches the largest value with $41.4$~N. 
A similar trend, but favoring the proposed approach, is observed for $F_{{ext}}^{{max}}$. Imp-ACT achieves the lowest mean peak force ($43.2$~N), followed by the low-stiffness baseline ($45$~N), while the high-stiffness baseline again has the largest value with $94.4$~N. When the completion time $t_c$ is analyzed, the high-stiffness baseline is fastest, finishing the task in 16.9~s, followed by Imp-ACT with 20.3~s and the low-stiffness baseline with 25.6~s. This indicates that Imp-ACT recovers part of the speed lost with compliant behavior, while still maintaining low interaction forces.

While the force and completion-time results reveal the expected trade-off between compliance and tracking performance, the most pronounced difference between the three approaches emerges in the smoothness of the contact interaction. \(F_{{ext}}^{{mean}}\) and \(F_{{ext}}^{{max}}\) characterize the overall load applied to the environment, but they do not capture rapid force fluctuations associated with oscillatory or stick--slip behavior. To quantify this aspect, we compute the band power of the vertical interaction force $F_z$ during contact. Band power $P_{vibr}$ is calculated as the trapezoidal integral of the Welch power spectral density of the mean-removed $F_z$ signal over the $8$--$40$ Hz frequency range. The force signal, sampled at $1$ kHz, may contain high-frequency measurement noise unrelated to the physical contact dynamics. Restricting the analysis to the selected band therefore focuses the metric on vibrations associated with contact chatter while reducing sensitivity to such noise. Imp-ACT achieves a band power of only $0.6$ N$^2$, compared with $17.7$ N$^2$ for the low-stiffness baseline and $109.6$ N$^2$ for the high-stiffness baseline, corresponding to reductions of approximately $29\times$ and $180\times$, respectively.
 We hypothesize that this reduction results from the directional stiffness adaptation. Wiping benefits from higher stiffness along the tangential direction to preserve motion against friction, while excessive normal stiffness can amplify contact disturbances. By increasing stiffness along the direction of motion while keeping the orthogonal directions compliant, Imp-ACT can preserve tangential tracking without unnecessarily increasing the normal interaction load.

\subsection{Plug insertion}

Plug insertion requires both accurate alignment and sufficient force along the insertion direction to overcome socket resistance. In this task, Imp-ACT achieves the highest success rate with \(76\%\), followed by the high-stiffness baseline with \(72\%\), and the low-stiffness baseline with \(60\%\) (see Fig.~\ref{fig:contact-metrics}). 

The proposed approach also results in the lowest interaction forces, with $F_{{ext}}^{{mean}}=12.3$ N and $F_{{ext}}^{{max}}=23.0$ N, compared with $13.8$ N and $31.3$ N for the low-stiffness baseline, and $20.3$ N and $38.4$ N for the high-stiffness baseline.
However, the magnitude of the total interaction force alone does not fully characterize insertion quality. During insertion, force along the insertion axis is expected and necessary to push the plug against the socket resistance, whereas forces orthogonal to this direction mainly arise from misalignment and contact with the socket walls. We therefore additionally evaluate the transverse interaction force ($F_{{ext}}^{{cross}}=\sqrt{F_x^2+F_y^2}$), which quantifies lateral loading associated with misalignment and contact with the socket walls. Imp-ACT reduces this force to $4.4$ N, compared with $7.8$ N for the low-stiffness baseline and $11.9$ N for the high-stiffness baseline, corresponding to a $43\%$ reduction relative to the better baseline. This result is consistent with the anisotropic adaptation mechanism, where higher stiffness along the insertion direction facilitates plug insertion, while orthogonal compliance allows residual misalignment to be accommodated with lower lateral loading.

Alongside these improvements in contact behavior, $t_c$ remains comparable across conditions ($15.8$ s for Imp-ACT, $15.1$ s for the high-stiffness baseline, and $18.2$ s for the low-stiffness baseline). These results indicate that Imp-ACT reduces interaction forces while maintaining a completion time similar to the faster baseline.


\section{Conclusion}

This work introduced Imp-ACT, a methodologically grounded approach to integrate stiffness modulation into demonstration collection for imitation learning. The framework preserves an intuitive, kinematics-based teleoperation interface in which the operator commands the desired motion while the self-tuning module adjusts stiffness online. Adaptive impedance is therefore part of the demonstration, without requiring explicit stiffness commands from the operator or targets assigned after data collection. This allows the policy to learn from demonstrations in which compliance is already present. The wiping and plug insertion experiments highlight the benefit of adapting stiffness to the directional demands of the task, supporting motion along the intended direction while preserving orthogonal compliance to limit transverse interaction forces and accommodate environmental constraints. Future work could extend the stiffness adaptation law to rotational degrees of freedom, enabling both translational and rotational compliance to be modulated and recorded during teleoperation. A further direction is to incorporate a passivity filter to regulate energy injection associated with stiffness variations and establish conditions for stable robot interaction with the environment. The data-collection principle is policy-agnostic in spirit, motivating future work on integrating stiffness modulation into a broader range of visual-action policy pipelines such as Diffusion Policy or more general foundation models like VLAs.


\section*{Acknowledgment}
This paper was supported by the European Union Horizon Projects TORNADO (Grant GA 101189557) and by the Italian Ministry of University and Research (MUR) under the Fondo Italiano per la Scienza (FIS), call FIS 3, project EPIC with code FIS-2024-02654.
The authors thank Sebastian Schleisner Hjorth for helpful discussions and the HRII$^2$ Lab for providing equipment and facilities. The authors also acknowledge IIT and its HPC Team for their support and assistance with the IIT High Performance Computing infrastructure.


\bibliographystyle{IEEEtran}
\bibliography{biblio}

\end{document}